\documentclass[journal,twoside,web]{ieeecolor}
\usepackage{generic}
\usepackage{amsmath,amssymb,amsfonts}
\usepackage{algorithmic}
\usepackage{graphicx}
\usepackage{textcomp}

\usepackage{multirow}
\usepackage[style=ieee]{biblatex}
\usepackage{xr}
\makeatletter
\newcommand*{\addFileDependency}[1]{
  \typeout{(#1)}
  \@addtofilelist{#1}
  \IfFileExists{#1}{}{\typeout{No file #1.}}
}
\makeatother

\newcommand*{\myexternaldocument}[1]{%
    \externaldocument{#1}%
    \addFileDependency{#1.tex}%
    \addFileDependency{#1.aux}%
}
\myexternaldocument{SI}

\def\BibTeX{{\rm B\kern-.05em{\sc i\kern-.025em b}\kern-.08em
    T\kern-.1667em\lower.7ex\hbox{E}\kern-.125emX}}
\begin{document}
\title{On Merging Feature Engineering and Deep Learning for Diagnosis, Risk-Prediction and Age Estimation Based on the 12-Lead ECG}
\author{Eran Zvuloni, Jesse Read, Antônio H. Ribeiro, Antonio Luiz P. Ribeiro and Joachim A. Behar
\thanks{This paragraph of the first footnote will contain the date on 
which you submitted your paper for review. It will also contain support 
information, including sponsor and financial support acknowledgment. For 
example, ``This work was supported in part by the U.S. Department of 
Commerce under Grant BS123456.'' }
\thanks{E. Zvuloni and J. A. Behar are with the Faculty of Biomedical Engineering, Technion-IIT, Haifa, Israel (e-mail: jbehar@technion.ac.il). }
\thanks{J. Read is with the DaSciM team of the Data Analytics and Machine Learning pole of the Computer Science Laboratory (LIX) at École Polytechnique, Institut Polytechnique de Paris.}
\thanks{A. H. Ribeiro is with the Department of Information Technology, Uppsala University, Uppsala, Sweden.}
\thanks{A. L. P. Ribeiro is with the Department of Internal Medicine, Faculdade de Medicina, Universidade Federal de Minas Gerais, Belo Horizonte, Brazil.}
}

\maketitle
\hfill \break
\begin{abstract}
\textit{Objective:} Machine learning techniques have been used extensively for 12-lead electrocardiogram (ECG) analysis. For physiological time series, deep learning (DL) superiority to feature engineering (FE) approaches based on domain knowledge is still an open question. Moreover, it remains unclear whether combining DL with FE may improve performance. \textit{Methods:} We considered three tasks intending to address these research gaps: cardiac arrhythmia diagnosis (multiclass-multilabel classification), atrial fibrillation risk prediction (binary classification), and age estimation (regression). We used an overall dataset of 2.3M 12-lead ECG recordings to train the following models for each task: i) a random forest taking the FE as input was trained as a classical machine learning approach; ii) an end-to-end DL model; and iii) a merged model of FE+DL. \textit{Results:} FE yielded comparable results to DL while necessitating significantly less data for the two classification tasks and it was outperformed by DL for the regression task. For all tasks, merging FE with DL did not improve performance over DL alone. \textit{Conclusion:} We found that for traditional 12-lead ECG based diagnosis tasks DL did not yield a meaningful improvement over FE, while it improved significantly the nontraditional regression task. We also found that combining FE with DL did not improve over DL alone which suggests that the FE were redundant with the features learned by DL. \textit{Significance:} Our findings provides important recommendations on what machine learning strategy and data regime to chose with respect to the task at hand for the development of new machine learning models based on the 12-lead ECG.
\end{abstract}


\section{Introduction}
\label{sec:introduction}

Machine learning based cardiovascular disease management using 12-lead electrocardiogram (ECG) analysis has been studied extensively over the last two decade~\cite{Minchole2019MachineElectrocardiogram, Sahoo2020MachineSurvey, Hong2020OpportunitiesReview}. A reliable system capable of assisting physicians’ decision-making process may significantly improve diagnosis and consequently reduce healthcare costs~\cite{Ribeiro2020AutomaticNetwork}.

Deep learning (DL) is broadly accepted as an effective and suitable data-driven approach in performing feature extraction for classification or regression. In the computer vision field it is the dominant or even exclusive approach for many data-driven tasks~\cite{Voulodimos2018DeepReview}. This is thanks to the inherent properties of convolutional layers which make them very good feature extractors in natural images~\cite{Bogatskiy2020LorentzPhysics}. Nevertheless, in 1D physiological time-series, and thus for 12-lead ECG analysis, DL superiority to classical feature engineering (FE) based machine learning approaches is still an open question. It also remains to be elucidated whether combining FE with DL may yield better performance and thus whether FE provides complementary or redundant information over DL.

One attempt investigating the connection between how ECG features are extracted through DL versus FE was made by Attia \textit{et al.}~\cite{Attia2021DeepFeatures}. The authors trained a model providing sex and age estimations by analyzing 12-lead ECG signals. A neural network modeled the relationship between DL and FE features. Moreover, linear combinations of the DL features were used to represent the engineered ones. Both experiments provided partial explainability to the DL features. Yet, the features did not obtain a perfect match by being fully modeled or represented, which implied that some information gaps remained. Another work by Beer \textit{et al.}~\cite{Beer2020UsingSignals} focused on the task of atrial fibrillation (AF) binary classification from a single lead ECG input. The authors added the Hilbert-Schmidt independence criterion to their loss function, constraining their DL model to learn different features from those provided by FE. This experiment succeeded in generating novel DL features. Nevertheless, these were not valuable features for the AF classification task, obtaining poor scores when the classification was confined to them. Interestingly, this work demonstrated how DL could discover similar features to those engineered by human and that may have taken years to formalize.

Despite the progress in finding association between FE and DL features for ECG analysis, there is a lack of quantitative analysis with respect to the level of redundancy that exist between these two approaches in performing classification or regression learning tasks. Moreover, in the research of ~\cite{Beer2020UsingSignals, Attia2021DeepFeatures} the database sizes used were relatively small, ranging from 12,186 to 100,000 ECG recordings respectively. Thus, the added value of combing FE and DL in a big data regime remains unclear. We decided to focus our experiments on 12-lead ECG analysis because it is the most commonly used cardiac examination in clinical practice.

Herein, we seek to benchmark and evaluate the superiority of DL algorithms to classical FE-based machine learning approaches, together with investigating the value of merging FE with DL for 12-lead ECG analysis and in a big data regime. Three tasks were considered: 1) multiclass-multilabel classification of 6 cardiac arrhythmia diagnosis, 2) binary classification for AF risk prediction, and 3) regression for age estimation. The different experiments for each task (FE exclusive, DL exclusive, and merged FE+DL) utilize the same pipeline and FE-DL merging approach, while the model optimization is task-specific.

\section{Methods}
In the following section we describe the: A) dataset used, B) FE types and their setup, C) model pipeline and architecture, D) performance measures used together with the statistical analysis and E) learning curves elaboration.

\subsection{Dataset}
The recordings are part of the Telehealth Network of Minas Gerais (TNMG) database. All the recordings were 7 to 10 seconds long and were resampled to a sampling frequency of 400Hz. The ECG recordings were originally used by Ribeiro \textit{et al.} study~\cite{Ribeiro2020AutomaticNetwork}. Table~\ref{tabDATA} shows the exact dataset division with respect to the learning tasks and for the different classes considered in our experiments. For the arrhythmia diagnosis task, all the recordings with arrhythmia labels were used, and the test set consisted of the additional 827 recordings with labels manually annotated by three different electrocardiography-expert cardiologists, as used in the previous research of~\cite{Ribeiro2020AutomaticNetwork}. The six classes of arrhythmias considered were: first-degree atrioventricular block (1dAVb), right and left bundle branch block (RBBB, LBBB), sinus bradycardia (SB), AF and sinus tachycardia (ST). For the AF risk prediction task, a subset of the dataset was used. The complete experimental setting is described in Biton \textit{et al.}~\cite{Biton2021AtrialLearning}. Briefly, recordings classified as ``Future AF" were those patients with a baseline recording without an AF label and with a subsequent recording within 5 years with a positive AF diagnosis. These correspond to class $C_2$ in~\cite{Biton2021AtrialLearning}. Whereas ``Non-AF" patients (class $C_1$ ) were those with a baseline recording with no AF documented and no AF documented in any subsequent recording within 5 years in~\cite{Biton2021AtrialLearning}. For the age estimation task, the test set followed the work of Lima \textit{et al.}~\cite{Lima2021DeepPredictor}; thus, single recordings per patients of ages 16-85 were taken from the CODE-15\%~\cite{Ribeiro2021CODE-15:ECGs} dataset (a subset of the TNMG database). Other recordings with patient age available were used for the train and validation sets.

\begin{table}
\caption{Dataset and data splitting according to the three tasks}

\begin{tabular}{p{0.15\columnwidth}p{0.15\columnwidth}p{0.1\columnwidth}p{0.1\columnwidth}p{0.1\columnwidth}p{0.1\columnwidth}}
\hline\hline

\textbf{Task}                                  & \textbf{Class}                   & \textbf{Train}   & \textbf{Validation} & \textbf{Test}   & \textbf{Total}   \\\hline
\multirow{9}{*}{\parbox{1.5cm}{\textbf{Arrhythmia diagnosis}}} & NSR                     & 2,029,146 & 41,412      & 681    & 2,071,239 \\
                                      & 1dAVb                   & 25,555   & 522        & 20     & 26,097   \\
                                      & RBBB                    & 49,876   & 1,018       & 28     & 50,922   \\
                                      & LBBB                    & 28,998   & 592        & 25     & 29,615   \\
                                      & SB                      & 32,386   & 661        & 15     & 33,062   \\
                                      & AF                      & 32,550   & 665        & 11     & 33,226   \\
                                      & ST                      & 45,199   & 923        & 35     & 46,157   \\
                                      & Multi-label             & 22,203   & 454        & 12     & 22,669   \\
                                      & \textbf{Total}                   & \textbf{2,265,913} & \textbf{46,247}      & \textbf{827}    & \textbf{2,290,318}\\
\hline\hline
\multirow{3}{*}{\parbox{1.5cm}{\textbf{Risk\\prediction}}}        & No AF                   & 797,786  & 16,282      & 203,568 & 1,017,636 \\
                                      & Future AF & 8,737    & 178        & 2,162   & 11,077   \\
                                      & \textbf{Total}                   & \textbf{806,523}  & \textbf{16,460}      & \textbf{205,730} & \textbf{1,028,713} \\
\hline\hline
\textbf{Age   estimation}                      & Regression              & \textbf{1,852,130} & \textbf{115,759}      & \textbf{233,233}  & \textbf{2,312,160} \\
\hline\hline
\end{tabular}
\label{tabDATA}
\end{table}

\subsection{Feature engineering}
Three types of features were used. The first type consisted in 16 self reported clinical variables  (META; see supplementary information Table 1). As for engineered features, a total of 23 heart rate variability (HRV) features were used including those listed in Chocron \textit{et al.}~\cite{Chocron2021RemoteNetwork} as well as three “extended parabolic phase space mapping” features~\cite{Moharreri2014ExtendedSignal}. Moreover, a mean signal quality index called bSQI~\cite{Behar2013ECGReduction} was computed as an additional HRV feature. The third set consisted in 22 morphological (MOR) features~\cite{9662857}. HRV and MOR features were computed for each lead, ending up with an overall of 557 engineered features per 12-lead ECG recording. For the cardiac arrhythmia diagnosis task, the age and sex were only available for the patients included in the expert reviewed test set and thus these were discarded, leading to 543 features per example. All the features were used for the AF risk prediction task. For the age estimation task, 556 features were available since the age was used as the target label and thus not included in the feature set.

\subsection{Model pipeline}

\begin{figure*}[t]
\centering
\centerline{\includegraphics[width=\textwidth]{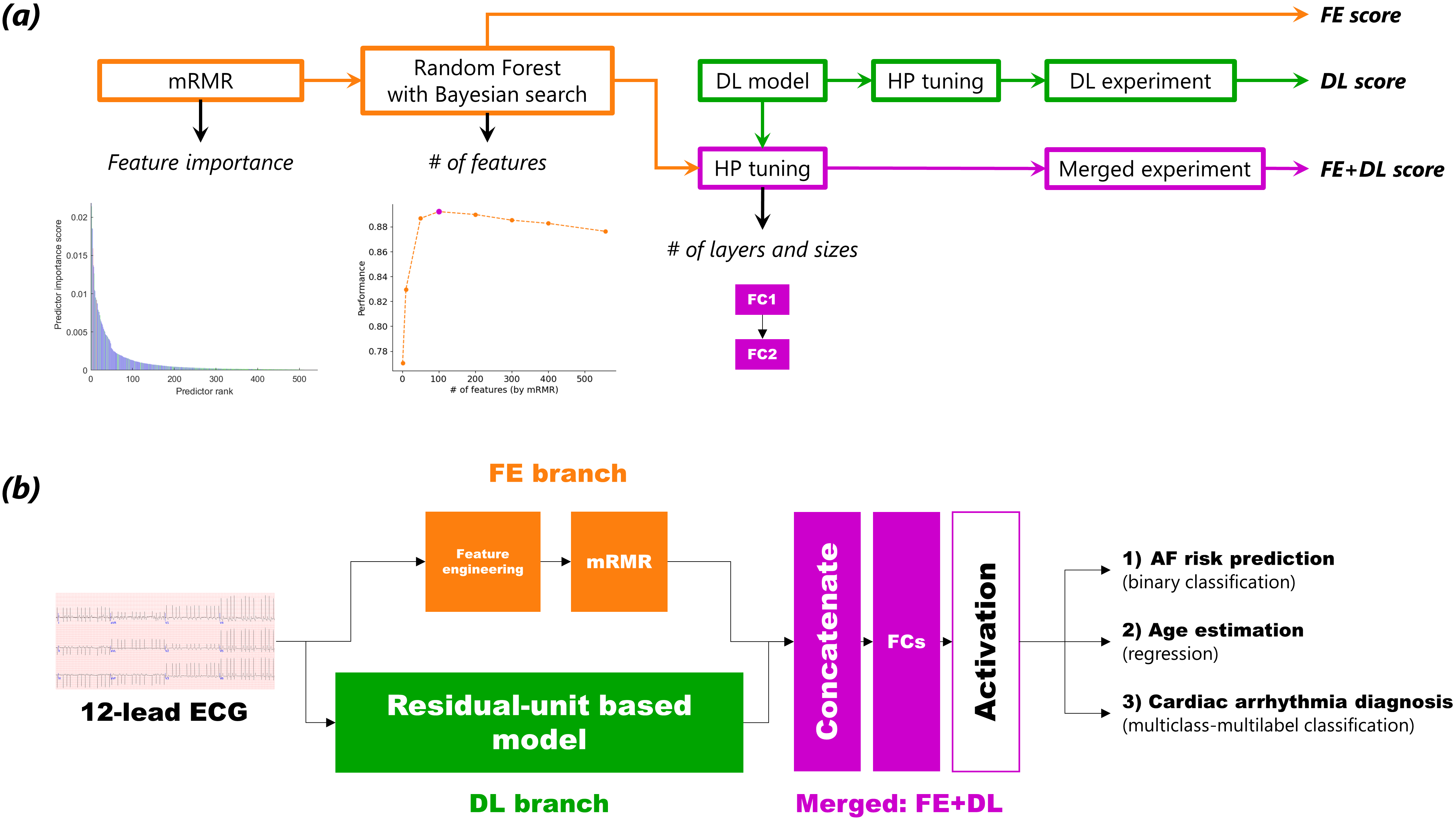}}
\caption{Model pipeline and overall architecture. Orange, green and purple components are associated with the feature engineering (FE), deep learning (DL) and merged (FE+DL) experiments, respectively. (a) Pipeline steps and feature inclusion in the merged model: 1. Engineered features importance is set by minimum redundancy maximum relevance (mRMR) algorithm. 2. Random forest (RF) classifier or regressor is trained and optimized by Bayesian search with altering number of features prioritized by the first step results. Best number of features is used for next steps and best RF model performance is taken as the FE score. 3. DL model hyperparameters (HP) are tuned. Then the DL experiment is conducted to obtain the DL score 4. HP tuning with merging the selected features and the DL model, including optimizing the fully connected (FC) layers after the FE and DL feature concatenation. 5. The merged experiment is conducted to obtain the FE+DL scores. (b) Overall model architecture design: a 12-lead ECG is the input for both branches. In the FE branch (orange) the signal is analyzed with feature engineering methods to extract features. Then, some of the features are selected with mRMR as explained in a. The DL branch (green) consists of deep neural network architecture taken from \cite{Ribeiro2020AutomaticNetwork}. Features from both branches are merged by concatenation in the purple head. These together are classified with FC layers and an activation that is altered according to one of the three output tasks.}
\label{fig_pipe}
\end{figure*}

The dataset split was different for each task (Table~\ref{tabDATA}). For the cardiac arrhythmia diagnosis task, the data were stratified according to the six different arrhythmias and split in 98\% train and 2\% validation similar to the work of Ribeiro \textit{et al.}~\cite{Ribeiro2020AutomaticNetwork}. The additional 827 recordings obtained by the expert consensus were used as the test set~\cite{Ribeiro2020AutomaticNetwork}. For the risk prediction task, the split was a 80\% train and 20\% test, similar to our previous work in Biton~et~al.~\cite{Biton2021AtrialLearning}. Then, 2\% of the examples taken from the train set split were used as the validation set. For the age estimation task, train and validation sets were split in a ratio of 80 to 5, as in Lima \textit{et al.}~\cite{Lima2021DeepPredictor}, while considering the recordings taken from the CODE-15\%~\cite{Ribeiro2021CODE-15:ECGs} as the test set. 

Experiments for the three tasks were conducted with the same pipeline consisting of the following steps (Fig. \ref{fig_pipe}a): (i)~Setting the engineered features importance using the minimum redundancy maximum relevance (mRMR) algorithm \cite{Ding2005MinimumData}. (ii)~Obtaining the FE performance score from a classical machine learning model by training a random forest (RF) classifier or regressor and taking FE as input. The number of selected features was set at this step to be later used as input to the concatenation layer (purple in Fig. \ref{fig_pipe}b) in the merged FE+DL experiments. (iii)~Hyperparameter tuning, including the fully connected (FC) layer adjustment added after the concatenation layer. (iv)~Running the DL and FE+DL experiments by training the models in a separated DL branch and in a merged FE+DL fashions to obtain their scores.

\subsubsection{Feature importance with mRMR}
mRMR algorithm~\cite{Ding2005MinimumData} from MATLAB R2020b (Mathworks) built-in functions was used. The mRMR was applied on the training set with respect to the relevant labels according to the task. For the diagnosis task, a multiclass-multilabel problem, we had to consider all the different labels as targets. Thus, we first treated the feature selection as multiple binary classification problems and applied mRMR with respect to each of the six labels (i.e., arrhythmias), obtaining six different rankings. Then, we used a union approach, i.e., when we set a number of selected features, there was at least that minimum number of features adequate to each of the labels in the final unified selected features. For example, by selecting 352 features to use, we in practice selected all 543 features of the diagnosis task. This is aligned with the approach used in the DL branch according to \cite{Ribeiro2020AutomaticNetwork}, as classifying the six arrhythmias was executed using a sigmoid activation function in the final layer. Thus, it addressed the task as a multiclass-multilabel problem, and the probabilities were obtained for each label independently with the others. The mRMR was applied with respect to the labels of the other two tasks as well: future AF labels as target for the AF risk prediction task to rank the 557 features (binary classification), and the patients' age for the age estimation task to rank the 556 features (regression). See supplementary information Fig. 1, 2, 3 for the mRMR results.

\subsubsection{Model training and hyperparameter tuning}
RF classifier and regressor of the scikit-learn 1.0.2 \cite{Pedregosa2011Scikit-learn:Python} Python package were used to perform classification and regression tasks with the engineered features. Hyperparameter tuning was performed with respect to the validation set performance. The optimized performance measure changed according to the task: harmonic mean of the positive predictive value (PPV) and sensitivity (Se), i.e., the F1-score, area under the receiver operating characteristic curve (AUROC) and negative mean absolute error (MAE), for the arrhythmia diagnosis, risk prediction, and age estimation tasks, respectively. The number of trees, maximum depth, split quality criterion and minimum number of samples at a leaf node were searched in a Bayesian optimization (scikit-optimize 0.9.0) together with the number of features (see supplementary information Table 2 for the selected parameters). In the classification tasks, class weights were applied to compensate the class-imbalance in the training sets. The least number of features yielding a performance plateau over the validation score was selected for the later FE+DL experiment in each task. In this way it was expected to reach a trade-off between the model degrees of freedom and its performance.

The neural network models were implemented using TensorFlow 2.5. The learning rate was set to $10^{-3}$ and was reduced by a factor of $10$ when the validation performance stopped improving for five consecutive epochs, similarly to \cite{Ribeiro2020AutomaticNetwork}. This allowed faster convergence in the initial epochs and finer refinements with smaller step sizes near an optimum. Binary cross-entropy was used as the loss function for the cardiac arrhythmia diagnosis and AF risk prediction classification tasks, whereas the MAE was the loss function used for the age estimation regression task. For all tasks, a batch size of 256 was used. Moreover, class-imbalance was corrected by assigning class weights according to their presence in the prediction and diagnosis tasks training sets. The number and size of the FC layers in the purple merging head (Fig. \ref{fig_pipe}b) were tuned as hyperparameters. 

\subsubsection{DL branch}
The DL branch was based on the model reported by Ribeiro \textit{et al.}~\cite{Ribeiro2020AutomaticNetwork}, and later used by Lima \textit{et al.}~\cite{Lima2021DeepPredictor} and Biton \textit{et al.}~\cite{Biton2021AtrialLearning}. Thus, it was taken as a benchmark model for our 12-lead ECG data. Briefly, 4 residual blocks extract the deep features, and end with a final classifying FC layer.

\subsubsection{Merged FE+DL model architecture}
Fig. \ref{fig_pipe}b shows the general model architecture including the FE branch (orange), the DL branch (green) and the merging head (purple) of the FE+DL experiments. The merging head included a concatenation layer which combined between the output neurons of both branches. These was followed by one or more FC layers creating the final network output. The final activation was task-based: sigmoid for classifications and no activation (linear) for the regression.

\subsection{Statistical analysis}
\subsubsection{Performance measures}
for the arrhythmia diagnosis task, the area under the precision recall curve (AUPRC) was used, as in~\cite{Ribeiro2020AutomaticNetwork}. We also reported the F1-score by computing the maximum score obtained when considering all thresholds used to build a precision recall curve. For the AF risk prediction task, the AUROC was used as in~\cite{Biton2021AtrialLearning, Kashou2020AProgram}. Last, for the age estimation task, a regression problem, the coefficient of determination ($R^2$) was computed to characterize the linear fitting between the input and output ages as in~\cite{Attia2019AgeECGs, Lima2021DeepPredictor}. The MAE which served as loss function was reported as well to describe the estimation error. AUPRC, AUROC and MAE were measured on the validation sets during the model training of the arrhythmia diagnosis, AF risk prediction and age estimation, respectively. These were used to track the deep model performances over the validation sets, and to select the best checkpoints for evaluation over the test sets. To compute the final AUPRC, the precision recall curve was processed by an interpolation and a median filter, followed by the trapezoid technique.

\subsubsection{Confidence intervals}
For each task, the confidence intervals were evaluated for the three experiments (i.e., FE, DL, FE+DL) using bootstrapping as in  \cite{Biton2021AtrialLearning}. Hence, the performances were computed on 1000 portions taken from the test set by randomly permuting its recordings and selecting only 80\% in each iteration. Then, a student's t-test was computed on each of the three pairs (i.e., FE vs DL, FE vs FE+DL and DL vs FE+DL). We defined the significance cut-off as $p_{value}<0.01$, which determined if the experiment results were significantly distinguishable.

\subsection{Learning curves}
Learning curves were produced to investigate how the number of ECG recordings used for training was affecting the performance of the different models. In particular, it could enable to assess how many ECG recordings were needed for DL performance to reach FE performance. For all the three tasks, these learning curves were produced for all experiments (i.e., FE, DL and FE+DL). The original hyperparameters set found when using the full training set were used in producing these curves as well.

\section{Results}
\subsection{Cardiac arrhythmia diagnosis}
\begin{figure}[h]
\centerline{\includegraphics[width=\columnwidth]{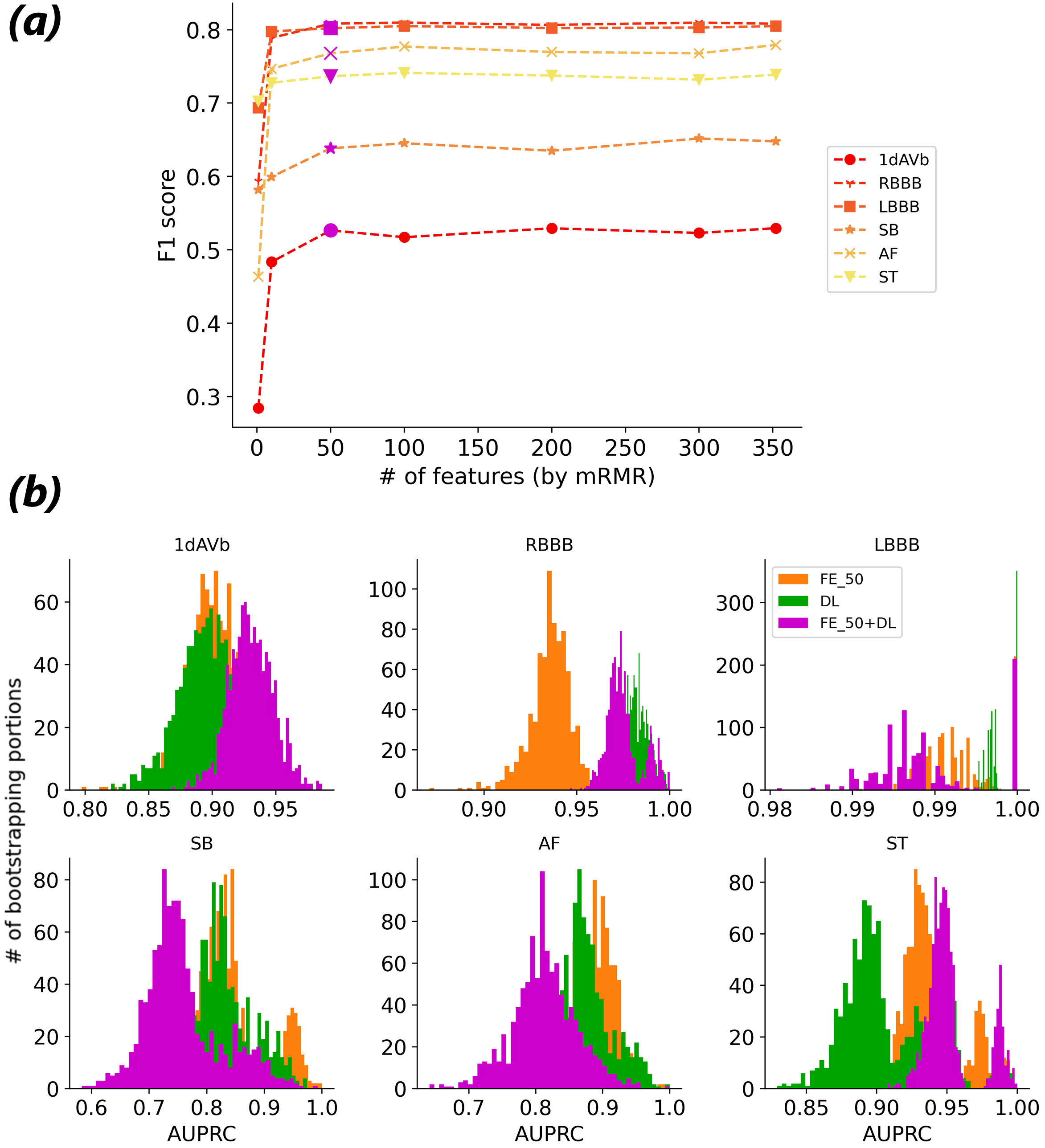}}
\caption{Arrhythmia diagnosis experiments. (a) Feature selection by minimum redundancy maximum relevance (mRMR) algorithm for the arrhythmia diagnosis task. Every line shows a separate random forest binary classification experiment based on different arrhythmia labels. The F1-score is reported on the validation set. 50 features (purple points) were selected for the feature engineering with deep learning (FE+DL) experiments, to include the features after all the classifiers reaching their plateau. (b) Bootstrapping over the test set results for each of the arrhythmias in each experiment. Orange, green and purple are the samples extracted from the test set results for the FE, DL and FE+DL experiments, respectively. The x-axis represents the area under the precision recall curve (AUPRC) score.}
\label{figDIAG1}
\end{figure}

The FE performance was reported for each arrhythmia individually (Fig. \ref{figDIAG1}a). All binary classification performances reached their plateau after selecting 50 features; therefore, for the FE+DL experiment we selected 50 features to include all the significant ones considering all labels. Since we worked with the union approach, this was effectively including 214 features to be used in the FE+DL experiment. For this task there was no need for additional FC layers between the concatenation and the final layers.

\begin{figure}[h]
\centering
\centerline{\includegraphics[width=\columnwidth]{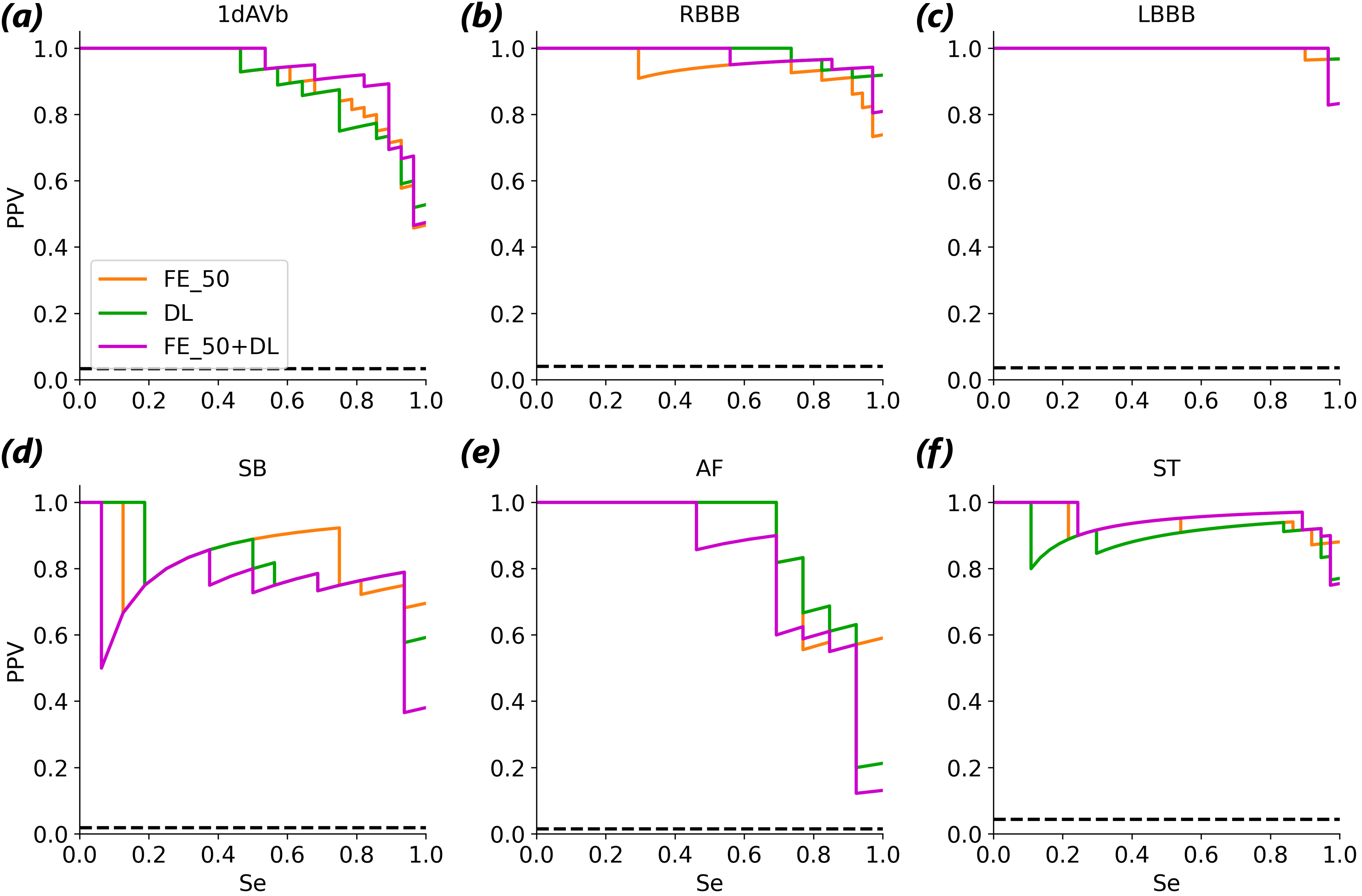}}
\caption{Precision-recall curves, i.e., positive predictive value (PPV) vs sensitivity (Se), for the arrhythmia diagnosis task. The results were obtained using the test set. Orange, green and purple lines shows feature engineering (FE), deep learning (DL) and merged branches experiment of FE together with DL, respectively. The black line resembles performance for a no-skill classifier (all predictions are positive). The corresponding area under the precision-recall curve (AUPRC) scores are summarized in Table \ref{tabDIAG}.}
\label{figDIAG2}
\end{figure}

The test set results for the three experiments are summarized in Table \ref{tabDIAG}. All experiments obtained similar results, while the FE mean AUPRC and the FE+DL mean F1 scores were the highest among the three, with $AUPRC_{FE}=0.92$ and $F1_{FE+DL}=0.90$. Nevertheless, different models performed better than others for individual arrhythmia diagnosis. For instance, for the RBBB arrhythmia, the DL experiment obtained best AUPRC performance with $AUPRC_{DL}=0.98$; yet, for AF, the FE experiment was the one to obtain highest AUPRC score with $AUPRC_{FE}=0.89$. Moreover, AF was the only case when the FE+DL experiment failed to deliver similar performance in both scores to FE or DL. The arrhythmia precision recall curves in Fig. \ref{figDIAG2} show the classifier behavior for different Se-PPV trade-offs. The bootstrapping results over the test set are shown in (Fig. \ref{figDIAG1}b). Though the performance scores were similar in most cases, all experiments were found significantly different with $p_{value}\ll0.01$, i.e., for FE vs DL; FE vs FE+DL and DL vs FE+DL.

\begin{table}[h]
\caption{Cardiac arrhythmia diagnosis task scores}
\begin{center}
\begin{tabular}{|c|c|c|c|c|}
\hline\hline
\textbf{Arrhythmia}             & \textbf{Test set score} & \textbf{FE}   & \textbf{DL}   & \textbf{FE+DL} \\\hline
\multirow{2}{*}{\textbf{1dAVb}} & AUPRC                   & 0.90          & 0.89          & \textbf{0.93}  \\
                                & F1                      & 0.83          & 0.81          & \textbf{0.88}  \\\hline
\multirow{2}{*}{\textbf{RBBB}}  & AUPRC                   & 0.94          & \textbf{0.98} & 0.97           \\
                                & F1                      & 0.91          & \textbf{0.96} & \textbf{0.96}  \\\hline
\multirow{2}{*}{\textbf{LBBB}}  & AUPRC                   & \textbf{1.00} & \textbf{1.00} & 0.99           \\
                                & F1                      & 0.97          & \textbf{0.98} & \textbf{0.98}  \\\hline
\multirow{2}{*}{\textbf{SB}}    & AUPRC                   & \textbf{0.84} & 0.83          & 0.75           \\
                                & F1                      & 0.83          & \textbf{0.86} & \textbf{0.86}  \\\hline
\multirow{2}{*}{\textbf{AF}}    & AUPRC                   & \textbf{0.89} & 0.87          & 0.81           \\
                                & F1                      & \textbf{0.82} & \textbf{0.82} & 0.78           \\\hline
\multirow{2}{*}{\textbf{ST}}    & AUPRC                   & 0.94          & 0.91          & \textbf{0.95}  \\
                                & F1                      & \textbf{0.94} & 0.93          & \textbf{0.94}  \\\hline\hline
\multirow{2}{*}{\textbf{Mean}}  & AUPRC                   & \textbf{0.92} & 0.91          & 0.90           \\
                                & F1                      & 0.88          & 0.89          & \textbf{0.90} \\\hline\hline
\end{tabular}
\end{center}
\label{tabDIAG}
\end{table}

\subsection{Risk prediction for atrial fibrillation}
\begin{figure}[h]
\centering
\centerline{\includegraphics[width=\columnwidth]{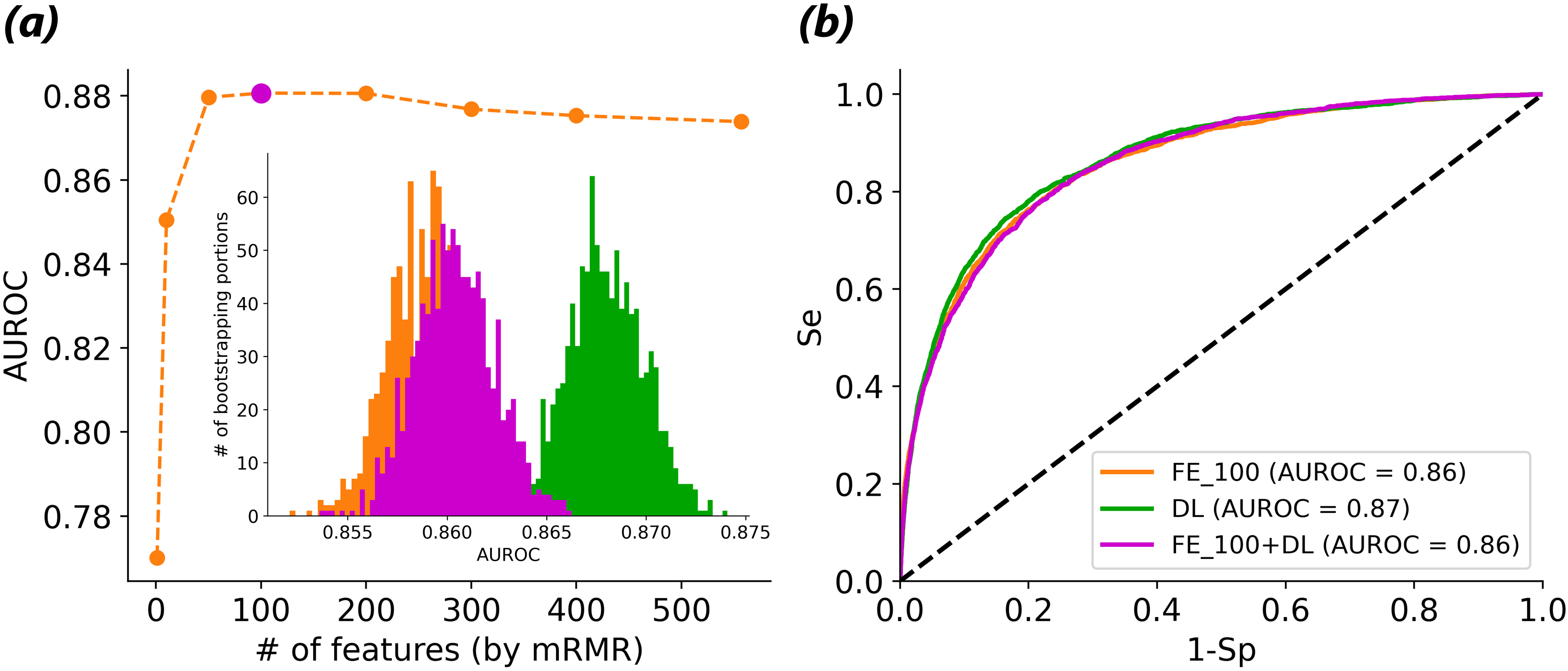}}
\caption{AF risk prediction task results. (a) Optimization of the number of features selected obtained by training a random forest (RF) classifier and reporting the area under the receiver operating characteristic curve (AUROC) performance over the validation set. Accordingly, 100 features were selected (purple point) as input to the merged experiments of feature engineering plus deep learning (FE+DL). Inset: Histograms after bootstrapping the three experiments (colors are according to the legend in (b) over the test set for statistical comparison. (b) ROC curve, i.e., sensitivity (Se) vs one minus specificity (Sp), and AUROC scores for the three experiments computed on the test set.}
\label{figPRED}
\end{figure}

Fig. \ref{figPRED} shows the results for the AF risk prediction task. The optimized validation score was obtained using 100 of the 557 features (Fig. \ref{figPRED}a) with importance set by the mRMR. With these, the RF classifier set a FE test score of $AUROC_{FE}=0.86$ (Fig. \ref{figPRED}b). After training the DL branch alone, there was a small improvement with $AUROC_{DL}=0.87$ compared to FE. For the FE+DL experiment, best performance for this task was obtained when adding another FC layer of 1000 neurons post concatenation and before the final layer. The FE+DL experiment reached a comparable AUROC score as the FE with $AUROC_{FE+DL}=0.86$. While bootstrapping over the test set, we observed that the DL performance increment was significant with respect to both the FE and FE+DL experiments (Fig. \ref{figPRED}a - inset). Yet this improvement was minor. The three AUROC curves (Fig. \ref{figPRED}b) were almost superimposed, suggesting similar classification performance in all experiments.

\subsection{Age estimation}
\begin{figure}[h]
\centering
\centerline{\includegraphics[width=\columnwidth]{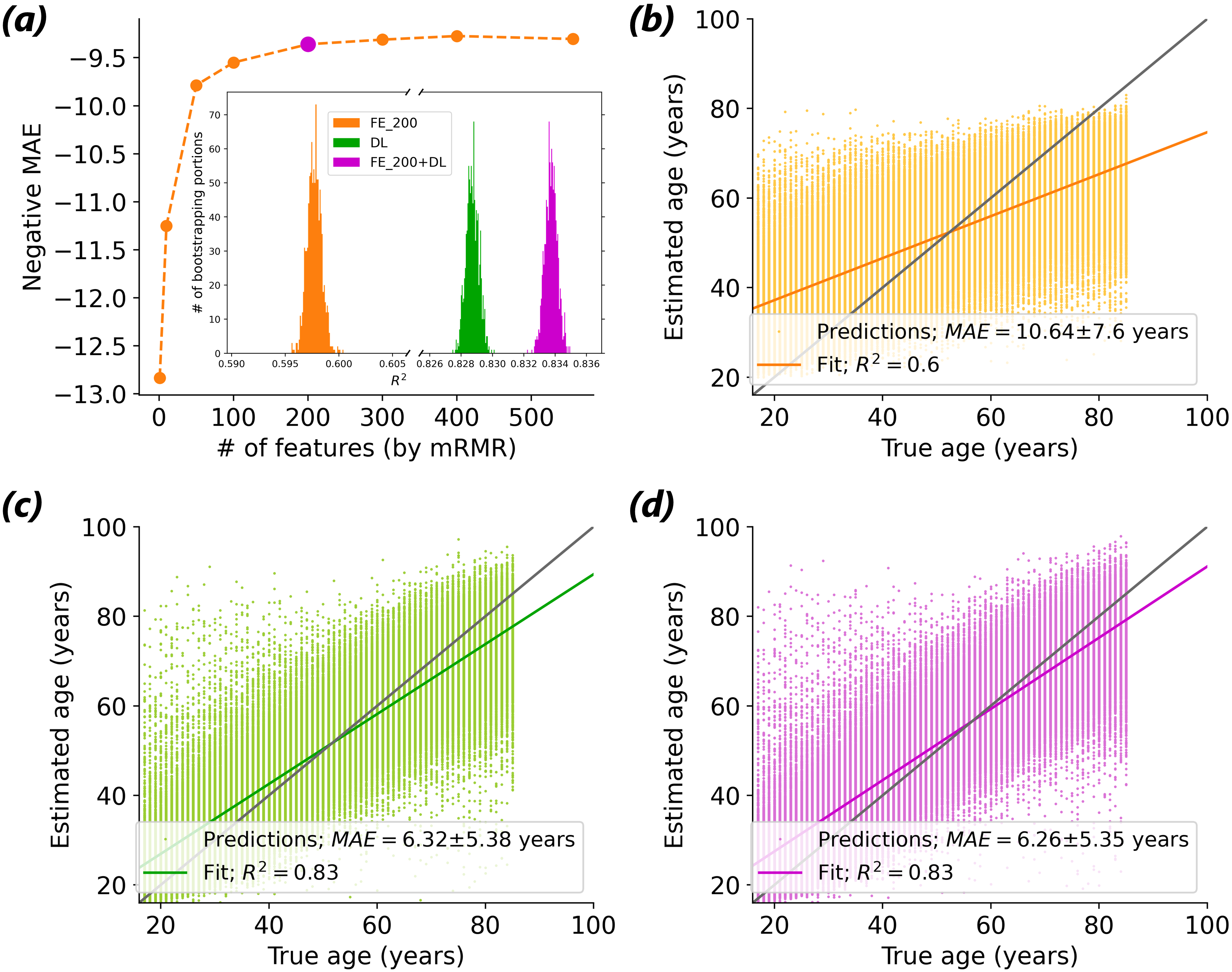}}
\caption{Age estimation task results. (a) Feature selection according to the random forest validation results. 200 features were selected, as performance reach plateau at this point (purple). Inset: Bootstrapping over the test set in the feature engineering (FE, orange), deep learning (DL, green) and FE+DL (purple) experiments. The inner axes are an enlargement of the outer one for a clearer view. (b-d) Test set results for the regression task for the three experiments: FE, DL and FE+DL in orange, green and purple, respectively. The results for each recording are displayed with their mean absolute error (MAE). Moreover, the overall linear fitting is displayed with the extracted $R^2$. The gray lines resemble identity between the estimated and true age ($y=x$).}
\label{figAGE}
\end{figure}

For the age estimation task, the RF regressor reached validation performance plateau starting with selecting 200 of the 556 features (Fig. \ref{figAGE}a). With these, the regressor reached test performance of $R^2_{FE}=0.60$ and $MAE_{FE}=10.64+7.6$ years (Fig. \ref{figAGE}b). Here, we did not find it useful to add FC layers between the concatenation and the final one. DL performed with $R^2_{DL}=0.83$; $MAE_{DL}=6.32+5.38$ years and DL+FE performed with $R^2_{FE+DL}=0.83$; $MAE_{FE+DL}=6.26+5.35$ (Fig. \ref{figAGE}c,d). The bootstrapping experiment in Fig. \ref{figAGE}a shows a significant ($p_{value}\ll0.01$) separation between DL and FE+DL although with a very small fold change in the mean performance.

\subsection{Altering the train set size}
\begin{figure*}[t]
\centering
\centerline{\includegraphics[width=\textwidth]{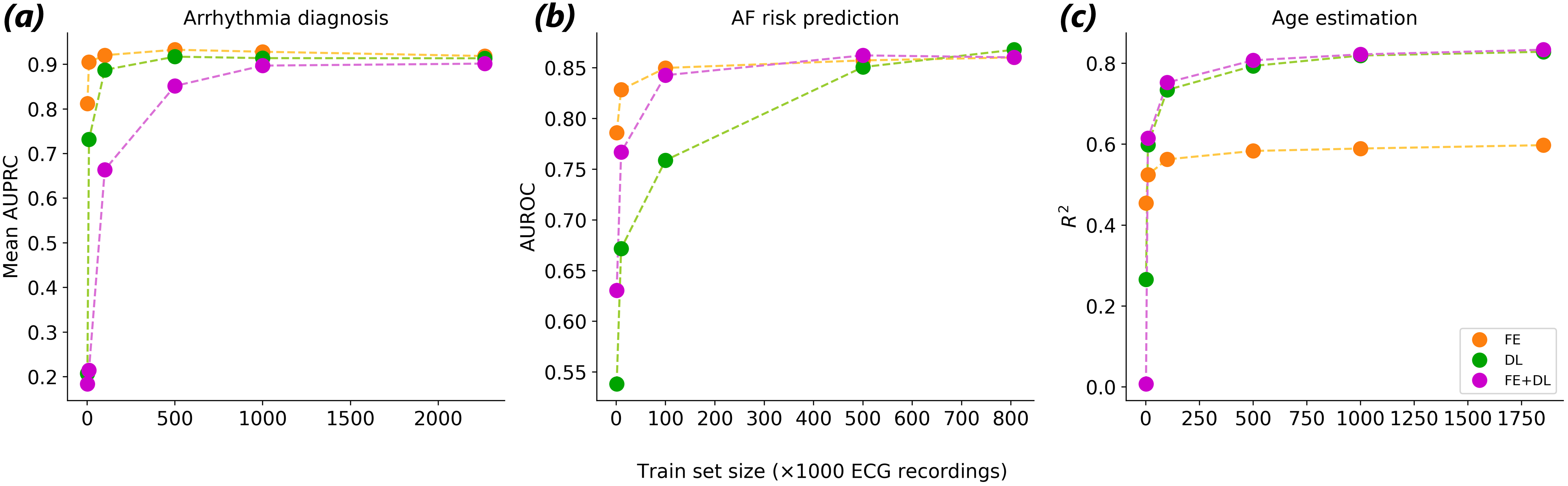}}
\caption{Learning curves. Altering train set size experiments for the three different tasks: (a) arrhythmia diagnosis; (b) atrial fibrillation (AF) risk prediction; and (c) age estimation. For all the tasks, smaller train set size experiments were added including $1k$, $10k$, $100k$, $500k$ or $1M$ ECG recordings, while the figures also show the maximum train set size that was used in the main experiments (Table \ref{tabDATA}). All the different experiments - feature engineering (FE, orange), deep learning (DL, green) and merged (FE+DL, purple) were conducted. The performance measures used here aligned with the main experiments: (a) the mean of the area under the precision recall curve (AUPRC) computed over the six arrhythmias; (b) area under the receiver operating characteristic curve (AUROC); and (c) the coefficient of determination ($R^2$). The horizontal axis is in units of thousand ECG recordings.}
\label{figALTER}
\end{figure*}

Fig. \ref{figALTER} shows the models performance as a function of the training set size. For the arrhythmias diagnosis task (Fig. \ref{figALTER}a), the performances of the FE and DL experiments were comparable for all different sizes, with a minor superiority for FE over DL data for small training sets (<500k examples). In this case the FE+DL experiment demonstrated slower curve than the others, which might be due to the model complexity. For the AF risk prediction case (Fig. \ref{figALTER}b), while FE performed better for small training sets, DL reached FE performance around $700k$ recordings. Since the maximal training set size was $806k$ recordings \ref{tabDATA}, it was not possible to observe whether DL would continue to improve with more examples. For the age estimation task (Fig. \ref{figALTER}c), DL performed systematically better than FE, with a clear advantage for all training set sizes. 

\section{Discussion}
For the cardiac arrhythmias diagnosis task, the FE and DL achieved a comparable mean score of $AUPRC_{FE}=0.92$, $F1_{FE}=0.88$ and $AUPRC_{DL}=0.91$, $F1_{DL}=0.89$, respectively. Thus, it is debatable what method is best, let alone when considering each individual arrhythmia. For instance, in the cases of 1dAVb and ST, the FE+DL experiments obtained higher scores than FE or DL. The DL experiment reproduced the work of Ribeiro \textit{et al.}~\cite{Ribeiro2020AutomaticNetwork}, and reached similar F1-scores as those originally reported.

For the AF risk prediction task, the results of the three models were comparable (Fig. \ref{figPRED}b) with DL yielding the best score with an $AUROC=0.87$. In our previous work by Biton \textit{et al.}~\cite{Biton2021AtrialLearning}, we predicted AF by exploiting a hybrid method, extracting latent features from a deep neural network which was previously trained for detecting AF presence and used these representation learning features as input to a RF classifier. This network achieved a score of $AUROC=0.82$ using the same dataset as in the present experiment, and thus is outperformed by the DL experiment conducted here. Therefore, our research demonstrated how the end-to-end DL method was a better option for this task. Nevertheless, when considering the FE integration, Biton \textit{et al.} results obtained $AUROC=0.91$ which outperforms our results. The higher performance of this hybrid model \textit{et al.} was due to the pretraining of the deep neural network on the additional $>1M$ examples that included AF examples, suggesting that this pretrained network had a significant impact in improving the performance. Thus, if pretraining is possible, it may represent an advantage of DL over FE.

For the age estimation task, DL significantly outperformed FE (Fig. \ref{figAGE}b vs Fig. \ref{figAGE}c). Only the DL was adequate to extract the necessary features relating to the patients age. This may be explained by the fact that engineered features are inherently drawn from domain knowledge about the relation between HRV or MOR features and cardiac conditions but not with patient demographics such as age. This suggests that for tasks with very little domain knowledge available, DL may be a better approach. Furthermore, our result achieved $R^2=0.83$ and thus were higher than in previous reports by Attia \textit{et al.}~\cite{Attia2019AgeECGs} and Lima \textit{et al.}~\cite{Lima2021DeepPredictor}, who reached scores of $R^2=0.7$ and $0.71$, respectively. The difference with Attia \textit{et al.}~\cite{Attia2019AgeECGs} may be explained by the larger dataset used in our experiments, $>2.3M$ versus $>770k$ recordings. Therefore our experiment leveraged the potential of DL to improve performance when trained on a large dataset (Fig. \ref{figALTER}c). In Lima \textit{et al.}~\cite{Lima2021DeepPredictor} the experiments were almost identical with respect to the data used. Nevertheless, the authors' model was originally tuned to predict mortality and thus included proportional weighting of the train set age groups to increase performance towards this specific target. Such strategy proved to be efficient with transferring the age estimation model to the mortality prediction task, but led to reduced performance for the age estimation task itself.

For the three tasks learning curves were produced in order to obtain insights on the minimal number of recordings necessary to reach the best performance. For the arrhythmia diagnosis task, the learning curves showed that in a small data regimen FE outperformed DL and that in a big data regiment there was no significant advantage for FE or DL (Fig. \ref{figALTER}a). For the AF risk prediction task we also observed that FE outperformed DL in a small data regimen. However, DL started to outperform FE at about $700k$ recordings (Fig. \ref{figALTER}b). There, the prediction task was more complex and less directly related to the original FE domain knowledge (as opposed to diagnosis), which possibly was what enabled the DL to outperform FE as the training dataset became larger. For the age estimation task the FE and the DL curves were distant from one another from very few examples and onward (Fig. \ref{figALTER}c) with DL constantly outperforming FE. We explain this by the fact that the age estimation task is far from the historical clinical tasks that traditional FE were designed for. Importantly, the altering train set size experiments demonstrated how certain tasks would actually require a relatively small dataset size to reach maximum performance. Specifically, the arrhythmia diagnosis could reach the maximized mean performance by using only $100k$ ECG recordings available with the FE approach, whereas a comparable performance with DL would have necessitated $500k$ recordings. For the AF risk prediction task, we could report with FE a high performance of an $AUROC=0.85$ by using only $100k$ ECGs, while the DL needed seven times more data to match this score. Finally, the age estimation task reached a plateau after $500k$ recordings regardless of the approach.  Overall, these learning curves provide valuable insights on whether DL provides an advantage (or not) over FE and the amount of data that is necessary to reach best performance.

Overall, the results showed that FE and DL performances were comparable for the two clinical tasks (diagnosis and risk prediction) while DL significantly outperformed FE ($R^2_{DL}=0.83$ vs $R^2_{FE}=0.60$) for the age estimation task. The HRV and MOR features were historically often engineered for the purpose of supporting diagnosis and were built according to medical scientific knowledge (domain knowledge) to serve this purpose. In that respect, we explain the comparable performance of FE and DL for such tasks while DL would perform better than FE for nontraditional clinical tasks such as age estimation from the ECG.

\textit{Limitations:} One important limitation was the limited test set size for the arrhythmia diagnosis task. The test set included only a small number of recordings for each arrhythmia that were reviewed by a board of cardiologists (Table \ref{tabDATA}). This limits the ability to assess the relative performance of each approach (FE, DL, FE+DL), and whether the fluctuations observed for each arrhythmia class would hold. Another important limitation is the number of recordings used in the train set of the AF risk prediction task, i.e. $806k$ examples against at least $1.8M$ for the other tasks, preventing us from further investigating whether the DL performance would continue increasing if more data was available (Fig. \ref{figALTER}b). An additional limitation is the explainability of the deep features. A key aspect in understanding the gap between FE and DL lies with comparing and explaining the information encoded in the deep features. Though this might not have been the scope of the present research, this should be further investigated.

\section{Conclusion}
We developed a unified model pipeline and trained it for three different and independent cardio-related tasks, utilizing both feature engineering and deep learning approaches as well as combining both. Experiments were conducted using a large dataset of over 2.3M 12-lead ECG recordings. For all tasks, cardiac arrhythmia diagnosis, atrial fibrillation risk prediction, and age estimation, the DL model reached similar or higher performance than the FE approach. However, DL improvement over FE was important only for the age estimation task. We explain this finding by the fact that age estimation is not a classical clinical task that the 12-lead ECG was historically used for. Consequently FE encompassing domain knowledge for this specific task was low. For the classification tasks, it was found to be difficult to justify a big data regime and the complexity of a DL model since the classical machine learning approach utilizing a much smaller dataset reached similar performances.

\section*{Acknowledgment}
EZ and JB acknowledge the support of the Technion EVPR Fund: Hittman Family Fund.

\printbibliography

@inproceedings{9662857,
    title = {{PhysioZoo ECG: Digital electrocardiography biomarkers to assess cardiac conduction}},
    year = {2021},
    booktitle = {2021 Computing in Cardiology (CinC)},
    author = {Gendelman, Sheina and Biton, Shany and Derman, Raphaël and Zvuloni, Eran and Levy, Jeremy and Lugassy, Snir and Alexandrovich, Alexandra and Behar, Joachim A},
    pages = {1--4},
    volume = {48}
}

@article{Kashou2020AProgram,
    title = {{A comprehensive artificial intelligence–enabled electrocardiogram interpretation program}},
    year = {2020},
    journal = {Cardiovascular Digital Health Journal},
    author = {Kashou, Anthony H. and Ko, Wei-Yin and Attia, Zachi I. and Cohen, Michal S. and Friedman, Paul A. and Noseworthy, Peter A.},
    number = {2},
    month = {9},
    pages = {62--70},
    volume = {1},
    publisher = {Elsevier Inc.}
}

@article{Attia2019AgeECGs,
    title = {{Age and Sex Estimation Using Artificial Intelligence From Standard 12-Lead ECGs}},
    year = {2019},
    journal = {Circulation: Arrhythmia and Electrophysiology},
    author = {Attia, Zachi I. and Friedman, Paul A. and Noseworthy, Peter A. and Lopez-Jimenez, Francisco and Ladewig, Dorothy J. and Satam, Gaurav and Pellikka, Patricia A. and Munger, Thomas M. and Asirvatham, Samuel J. and Scott, Christopher G. and Carter, Rickey E. and Kapa, Suraj},
    number = {9},
    month = {9},
    pages = {1--11},
    volume = {12}
}

@article{Biton2021AtrialLearning,
    title = {{Atrial fibrillation risk prediction from the 12-lead electrocardiogram using digital biomarkers and deep representation learning}},
    year = {2021},
    journal = {European Heart Journal - Digital Health},
    author = {Biton, Shany and Gendelman, Sheina and Ribeiro, Antônio H and Miana, Gabriela and Moreira, Carla and Ribeiro, Antonio Luiz P and Behar, Joachim A},
    month = {8}
}

@article{Ribeiro2020AutomaticNetwork,
    title = {{Automatic diagnosis of the 12-lead ECG using a deep neural network}},
    year = {2020},
    journal = {Nature Communications},
    author = {Ribeiro, Antônio H. and Ribeiro, Manoel Horta and Paix{\~{a}}o, Gabriela M. M. and Oliveira, Derick M. and Gomes, Paulo R. and Canazart, Jéssica A. and Ferreira, Milton P. S. and Andersson, Carl R. and Macfarlane, Peter W. and Meira, Wagner and Sch{\"{o}}n, Thomas B. and Ribeiro, Antonio Luiz P.},
    number = {1},
    month = {12},
    pages = {1760},
    volume = {11},
    publisher = {Springer US}
}

@article{Ribeiro2021CODE-15:ECGs,
    title = {{CODE-15{\%}: a large scale annotated dataset of 12-lead ECGs}},
    year = {2021},
    author = {Ribeiro, Antônio H. and Paixao, Gabriela M.M. and Lima, Emilly M. and Horta Ribeiro, Manoel and Pinto Filho, Marcelo M. and Gomes, Paulo R. and Oliveira, Derick M. and Meira Jr, Wagner and Schon, Thömas B and Ribeiro, Antonio Luiz P.},
    month = {6},
    url = {https://zenodo.org/record/4916206},
    doi = {10.5281/ZENODO.4916206}
}

@article{Voulodimos2018DeepReview,
    title = {{Deep Learning for Computer Vision: A Brief Review}},
    year = {2018},
    journal = {Computational Intelligence and Neuroscience},
    author = {Voulodimos, Athanasios and Doulamis, Nikolaos and Doulamis, Anastasios and Protopapadakis, Eftychios},
    pages = {1--13},
    volume = {2018}
}

@article{Lima2021DeepPredictor,
    title = {{Deep neural network-estimated electrocardiographic age as a mortality predictor}},
    year = {2021},
    journal = {Nature Communications},
    author = {Lima, Emilly M. and Ribeiro, Antônio H. and Paix{\~{a}}o, Gabriela M. M. and Ribeiro, Manoel Horta and Pinto-Filho, Marcelo M. and Gomes, Paulo R. and Oliveira, Derick M. and Sabino, Ester C. and Duncan, Bruce B. and Giatti, Luana and Barreto, Sandhi M. and Meira Jr, Wagner and Sch{\"{o}}n, Thomas B. and Ribeiro, Antonio Luiz P.},
    number = {1},
    month = {12},
    pages = {5117},
    volume = {12},
    publisher = {Springer US}
}

@article{Attia2021DeepFeatures,
    title = {{Deep neural networks learn by using human-selected electrocardiogram features and novel features}},
    year = {2021},
    journal = {European Heart Journal-Digital Health},
    author = {Attia, Zachi I and Lerman, Gilad and Friedman, Paul A},
    number = {3},
    pages = {446--455},
    volume = {2},
    publisher = {Oxford University Press}
}

@article{Behar2013ECGReduction,
    title = {{ECG signal quality during arrhythmia and its application to false alarm reduction}},
    year = {2013},
    journal = {IEEE transactions on biomedical engineering},
    author = {Behar, Joachim and Oster, Julien and Li, Qiao and Clifford, Gari D},
    number = {6},
    pages = {1660--1666},
    volume = {60},
    publisher = {IEEE}
}

@article{Moharreri2014ExtendedSignal,
    title = {{Extended Parabolic Phase Space Mapping (EPPSM): Novel quadratic function for representation of Heart Rate Variability signal}},
    year = {2014},
    journal = {Computing in Cardiology},
    author = {Moharreri, Sadaf and Rezaei, S. and Dabanloo, N. J. and Parvaneh, S.},
    pages = {417--420}
}

@article{Bogatskiy2020LorentzPhysics,
    title = {{Lorentz Group Equivariant Neural Network for Particle Physics}},
    year = {2020},
    journal = {37th International Conference on Machine Learning, ICML 2020},
    author = {Bogatskiy, Alexander and Anderson, Brandon and Offermann, Jan T. and Roussi, Marwah and Miller, David W. and Kondor, Risi},
    month = {6},
    pages = {969--979},
    volume = {PartF16814}
}

@article{Sahoo2020MachineSurvey,
    title = {{Machine Learning Approach to Detect Cardiac Arrhythmias in ECG Signals: A Survey}},
    year = {2020},
    journal = {IRBM},
    author = {Sahoo, S. and Dash, M. and Behera, S. and Sabut, S.},
    number = {4},
    month = {8},
    pages = {185--194},
    volume = {41},
    publisher = {Elsevier Masson SAS}
}

@article{Minchole2019MachineElectrocardiogram,
    title = {{Machine learning in the electrocardiogram}},
    year = {2019},
    journal = {Journal of Electrocardiology},
    author = {Minchol{\'{e}}, Ana and Camps, Julià and Lyon, Aurore and Rodr{\'{i}}guez, Blanca},
    month = {11},
    pages = {S61-S64},
    volume = {57},
    publisher = {Elsevier Inc.}
}

@article{Ding2005MinimumData,
    title = {{Minimum Redundancy Feature Selection from Microarray Gene Expression Data}},
    year = {2005},
    journal = {Journal of Bioinformatics and Computational Biology},
    author = {Ding, Chris and Peng, Hanchuan},
    number = {02},
    month = {4},
    pages = {185--205},
    volume = {03}
}

@article{Hong2020OpportunitiesReview,
    title = {{Opportunities and challenges of deep learning methods for electrocardiogram data: A systematic review}},
    year = {2020},
    journal = {Computers in Biology and Medicine},
    author = {Hong, Shenda and Zhou, Yuxi and Shang, Junyuan and Xiao, Cao and Sun, Jimeng},
    number = {December 2019},
    month = {7},
    pages = {103801},
    volume = {122},
    publisher = {Elsevier Ltd}
}

@article{Chocron2021RemoteNetwork,
    title = {{Remote Atrial Fibrillation Burden Estimation Using Deep Recurrent Neural Network}},
    year = {2021},
    journal = {IEEE Transactions on Biomedical Engineering},
    author = {Chocron, Armand and Oster, Julien and Biton, Shany and Mandel, Franck and Elbaz, Meyer and Zeevi, Yehoshua Y and Behar, Joachim A},
    number = {8},
    month = {8},
    pages = {2447--2455},
    volume = {68}
}

@article{Pedregosa2011Scikit-learn:Python,
    title = {{Scikit-learn: Machine learning in Python}},
    year = {2011},
    journal = {the Journal of machine Learning research},
    author = {Pedregosa, Fabian and Varoquaux, Gaël and Gramfort, Alexandre and Michel, Vincent and Thirion, Bertrand and Grisel, Olivier and Blondel, Mathieu and Prettenhofer, Peter and Weiss, Ron and Dubourg, Vincent},
    pages = {2825--2830},
    volume = {12},
    publisher = {JMLR. org}
}

@article{Beer2020UsingSignals,
    title = {{Using Deep Networks for Scientific Discovery in Physiological Signals}},
    year = {2020},
    author = {Beer, Tom and Eini-Porat, Bar and Goodfellow, Sebastian and Eytan, Danny and Shalit, Uri},
    month = {8},
    pages = {1--24}
}

\end{document}


\begin{titlepage}
   \begin{center}
       \vspace*{1cm}
        \large
       \textbf{Supplementary Information}

       \vspace{1cm}
       \Large
        \textbf{On merging feature engineering and deep learning for diagnosis, risk-prediction and age estimation based on the 12-lead ECG}
            
       \vspace{1.5cm}
        \normalsize
       Eran Zvuloni, Jesse Read, Antônio H. Ribeiro, Antonio Luiz P. Ribeiro and Joachim A. Behar
   \end{center}
\end{titlepage}

\tableofcontents

\newpage
\addcontentsline{toc}{section}{SI Table \ref{SI:tabMETA}: META features}
\begin{table}[h]
\centering
\begin{tabular}{ll}
\hline\hline
\textbf{\#} & \textbf{Feature name}                            \\\hline
1  & Age                                     \\
2  & Amiodarone                              \\
3  & Diuretics                               \\
4  & Sex                                     \\
5  & Pulmonary arterial hypertension         \\
6  & Chagas                                  \\
7  & Beta blockers                           \\
8  & Family coronary heart disease           \\
9  & Myocardial infarction                   \\
10 & Chronic   obstructive pulmonary disease \\
11 & Diabetes mellitus                       \\
12 & Obesity                                 \\
13 & Dyslipodemia                            \\
14 & Calcium blockers                        \\
15 & Smoking                                 \\
16 & Chronic kidney disease\\\hline\hline
\end{tabular}
\caption{Demographics and commodities (META) features. Electronic medical record (EMR) system data taken from the ECG records and used as features. 16 META features were extracted and used with the risk prediction and age estimation tasks. For the arrhythmia diagnosis task only age and sex were used.}
\label{SI:tabMETA}
\end{table}

\begin{figure}[h]
\includegraphics[width=\textwidth]{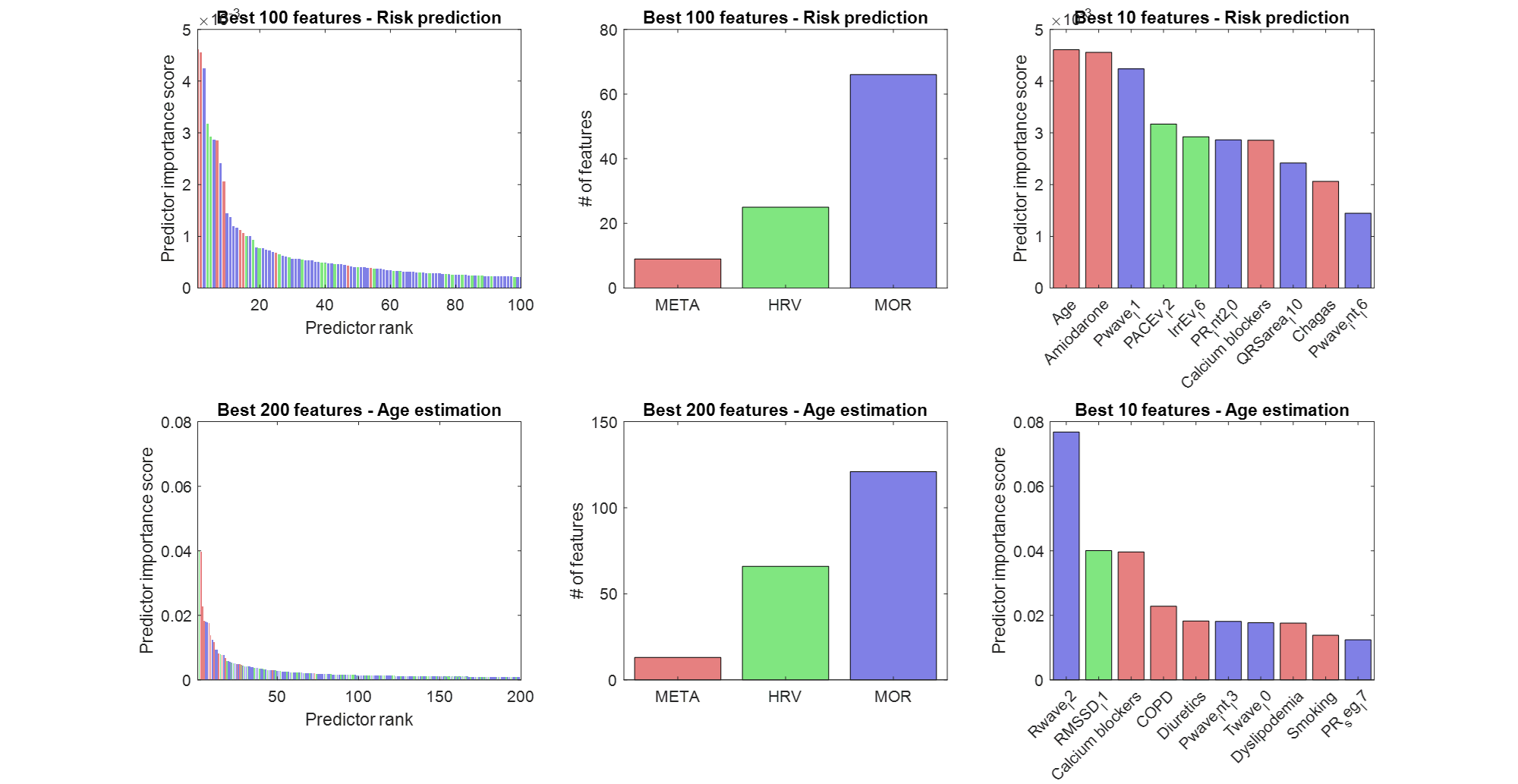}
\centering
\caption{Minimum redundancy maximum relevance (mRMR) algorithm~\cite{Ding2005MinimumData} results for the risk prediction and age estimation tasks. Results are shown for the number of features used in the final training, i.e., 100 features for the atrial fibrillation risk prediction task and 200 features for the age estimation task. Colors resemble different feature type: red, green and blue for demographics and commodities (META), heart rate variability (HRV) and morphological (MOR) features, respectively. The right figure shows the best 10 features with their names \cite{Chocron2021RemoteNetwork, 9662857, Moharreri2014ExtendedSignal}.}
\label{SI:fig_mRMR}
\end{figure}

\newpage
\addcontentsline{toc}{section}{SI Figure \ref{SI:fig_mRMR}: mRMR features - risk prediction and age estimation}

\begin{figure}[h]
\includegraphics[width=\textwidth]{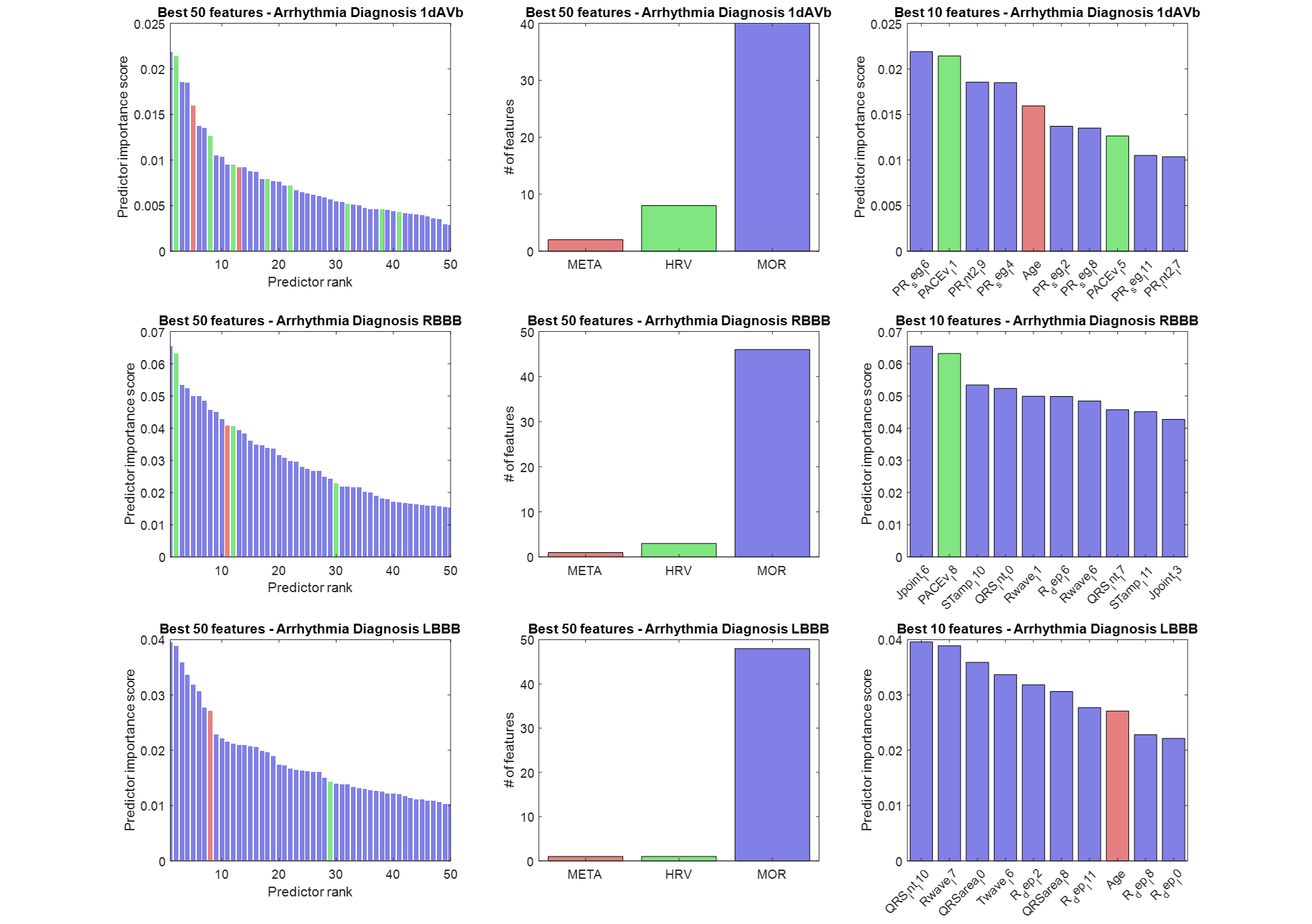}
\centering
\caption{Minimum redundancy maximum relevance (mRMR) algorithm results for three of the six arrhythmia in the diagnosis task: first-degree atrioventricular block (1dAVb) and right and left bundle branch block (RBBB, LBBB). A union of the best 50 features from all six arrhythmias was used for the final training. See SI Figure \ref{SI:fig_mRMR} for more information.}
\label{SI:fig_mRMR_DIAG}
\end{figure}

\newpage
\addcontentsline{toc}{section}{SI Figure \ref{SI:fig_mRMR_DIAG}: mRMR features - diagnosis: 1dAVb, RBBB and LBBB}

\begin{figure}[h]
\includegraphics[width=\textwidth]{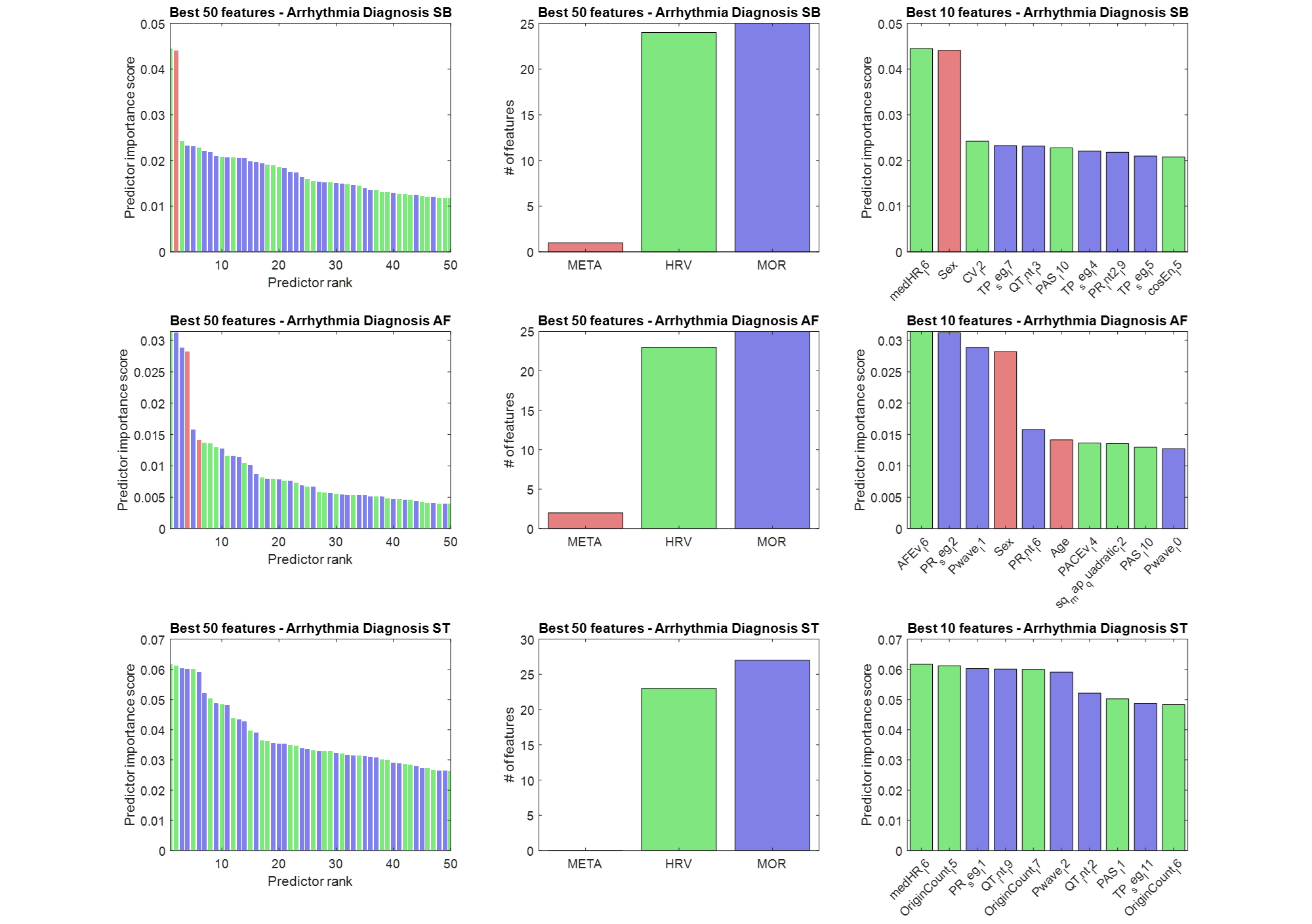}
\centering
\caption{Minimum redundancy maximum relevance (mRMR) algorithm results for three of the six arrhythmia in the diagnosis task: sinus bradycardia (SB), atrial fibrillation (AF) and sinus tachycardia (ST). A union of the best 50 features from all six arrhythmias was used for the final training. See SI Figure \ref{SI:fig_mRMR} for more information.}
\label{SI:fig_mRMR_DIAG2}
\end{figure}

\newpage
\addcontentsline{toc}{section}{SI Figure \ref{SI:fig_mRMR_DIAG2}: mRMR features - diagnosis: SB, AF and ST}

\newpage
\addcontentsline{toc}{section}{SI Table \ref{SI:tabBS_RF}: Bayesian search selected parameters}
\begin{table}[h]
\centering
\begin{tabular}{|m{2cm}|m{2.5cm}|m{1.5cm}|m{1.5cm}|m{2.5cm}|}
\hline\hline
                              & \textbf{Criterion} & \textbf{Max depth} & \textbf{Min samples leaf} & \textbf{n estimators} \\\hline 
\textbf{AF risk prediction}   & gini               & 100                & 30                        & 200                   \\\hline 
\textbf{Age estimation}       & squared error (fixed)      & 39                 & 1                         & 200                   \\\hline 
\textbf{Arrhythmia diagnosis} & entropy            & 28                 & 1                         & 38 \\\hline\hline                  
\end{tabular}
\caption{Bayesian search (BS) selected parameters. For the three tasks, the random forest max depth, min samples leaf and n estimators parameters were selected using BS by increasing the validation set performance. In the classification tasks (AF risk prediction and arrhythmia diagnosis) the criterion parameter was selected by the BS as well.}
\label{SI:tabBS_RF}
\end{table}

\newpage
\printbibliography